%% file: iccv_samcod.tex
\documentclass[10pt,twocolumn,letterpaper]{article}

\usepackage[pagenumbers]{iccv} 

\input{preamble}

\definecolor{iccvblue}{rgb}{0.21,0.49,0.74}
\usepackage[pagebackref,breaklinks,colorlinks,allcolors=iccvblue]{hyperref}

\def\paperID{7742} 
\def\confName{ICCV}
\def\confYear{2025}

\title{Improving SAM for Camouflaged Object Detection via Dual Stream Adapters}

\author{Jiaming Liu, {Linghe Kong\thanks{Corresponding author: Linghe Kong.} ,} Guihai Chen\\
	Department of Computer Science and Engineering, Shanghai Jiao Tong University\\
	Shanghai, China\\
	{\tt\small \{jmliu99, linghe.kong\}@sjtu.edu.cn, gchen@cs.sjtu.edu.cn}
}

\begin{document}
\maketitle

\begin{abstract}
	Segment anything model (SAM) has shown impressive general-purpose segmentation performance on natural images, but its performance on camouflaged object detection (COD) is unsatisfactory. In this paper, we propose SAM-COD that performs camouflaged object detection for RGB-D inputs. While keeping the SAM architecture intact, dual stream adapters are expanded on the image encoder to learn potential complementary information from RGB images and depth images, and fine-tune the mask decoder and its depth replica to perform dual-stream mask prediction. In practice, the dual stream adapters are embedded into the attention block of the image encoder in a parallel manner to facilitate the refinement and correction of the two types of image embeddings. To mitigate channel discrepancies arising from dual stream embeddings that do not directly interact with each other, we augment the association of dual stream embeddings using bidirectional knowledge distillation including a model distiller and a modal distiller. In addition, to predict the masks for RGB and depth attention maps, we hybridize the two types of image embeddings which are jointly learned with the prompt embeddings to update the initial prompt, and then feed them into the mask decoders to synchronize the consistency of image embeddings and prompt embeddings. Experimental results on four COD benchmarks show that our SAM-COD achieves excellent detection performance gains over SAM and achieves state-of-the-art results with a given fine-tuning paradigm. 
\end{abstract}

\section{Introduction}
SAM \cite{kirillov2023segment} is a visual foundation model (VFM) for promptable image segmentation with strong zero-shot capability and wide generalization. A fact is that SAM's ability to discriminate 2D images depends on the distribution coverage from the training data. Thus, a natural question arises: \textit{can SAM be directly extended to solve the task of segmenting camouflaged images for special scenarios?} Recent research \cite{chen2023sam,ji2024segment} shows that this is not feasible, as camouflaged images have similarities with natural images such as low contrast, which may even deceive the human eye \cite{stevens2009animal}.
\begin{figure}[t]
	\centering
	\includegraphics[width=\linewidth]{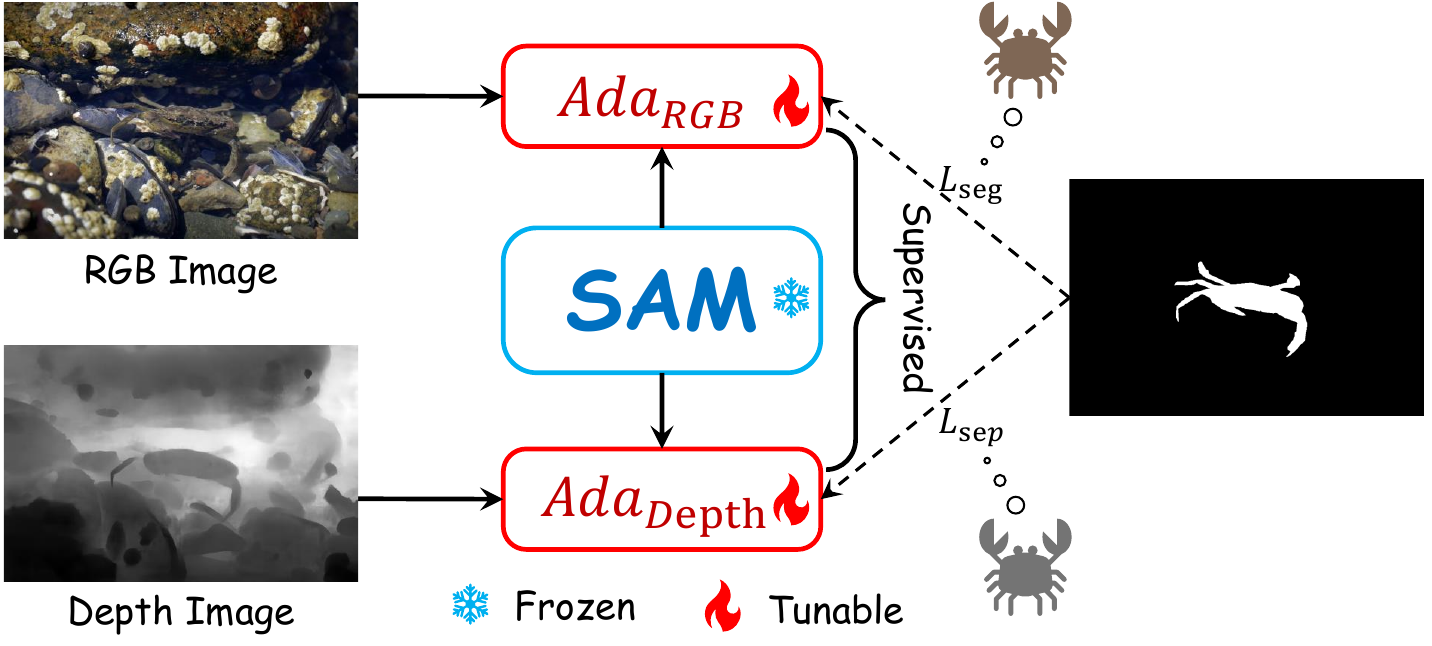}
	\vspace{-15pt}
	\caption{Illustration of the proposed SAM-COD. The two learnable adapters based on the SAM model are extended to act on paired input streams of RGB images and depth images, respectively. Subject to co-supervision, $Ada_{RGB}$ focuses on perceiving pixel semantics to segment objects and $Ada_{Depth}$ focuses on structural transformation to separate objects from the background.}
	\vspace{-10pt}
	\label{fig1}
\end{figure}

Based on the above observations, we further argue that fine-tuning is an important step in applying SAM to camouflaged images. We first analyze two key advantages associated with SAM: (1) The training dataset of SAM inherently contains a large number of backgrounds of camouflaged images, and the fixed parameters are sufficient to parse the visual environment. (2) After fine-tuning without changing the original weights, SAM still has strong generalization potential as a pre-trained large-scale VFM, which is crucial for effectively deploying models for COD in images with large inter-domain differences.

We then aim to redesign SAM to transfer it to a specific application taking into account domain knowledge. Recent studies on parameter-efficient adaptation methods have attempted to update pretrained parameters by specifying a small fraction of the parameters to be fine-tuned \cite{zaken2021bitfit,peng2024sam,paranjape2024adaptivesam} or merging lightweight adapters \cite{chen2024ma,iccv2023sam,wu2023medical}. The positive results have been achieved even with only a small number of parameter modifications or additions, which motivates us to seek an efficient version of the SAM adapter for camouflaged imaging. In contrast to previous research, our desired adapter needs to be applicable to both modalities, as RGB images and depth images can naturally complement obscure regions and merge valid object clues.

It is worth investigating how to design adapters with both modalities in mind — \textit{should we use fine-tuned networks with the same architecture or different ones?} Inspired by the teacher and student networks in DSAM \cite{yu2024exploring}, we believe that different modal inputs should be extracted with features in different ways in order to perform complementary functions. To this end, we design two adapters to handle RGB-D inputs. Considering the similar semantic and depth differences between objects and backgrounds in the COD, we aim to capture the high-frequency semantic features of the object from the RGB perspective to determine the transformed salient regions in the whole image, and then capture the geometric structural features of the object from the depth perspective to highlight the whole salient object from the background, as shown in Figure \ref{fig1}. Empirically, this semantic-structure fusion learning approach is well suited for detecting objects from various camouflaged images.

Specifically, we introduce the RGB adapter and the deep adapter as a combination of linear layer networks with frequency separation, using down-projection and up-projection linear layers embedded at both ends of the high-frequency feature extractor. The difference is that two frequency separation operations are performed under different specific filtering operators. Depending on the depth of the ViT network \cite{dosovitskiy2020image} and the number of attention blocks in the SAM, the extracted features of RGB images and depth images can be updated multiple times, resulting in the RGB embedding and depth embedding that contains rich details of camouflaged objects. In addition, to further update the initial box prompt, we synchronously use the RGB embedding and the depth embedding to blend with the prompt embedding, and perform channel-wise convolution operations in a multi-kernel manner, which makes the dense prompt embeddings close to the representations of the image embeddings on the feature channel. These encoder modules are jointly fine-tuned to transform generic visual perception models into specific models suitable for the COD task.

Moreover, since camouflaged objects have small size, irregular shape, and low contrast, we choose to continue activating two mask decoders to generate mask predictions with refined prompt embeddings and image embeddings. Considering that the source-free depth images may suffer from noise, we adapt inter-modal knowledge distillation, regarding the RGB embedding and depth embedding as the teacher and student, respectively. Our intuition is that since RGB embeddings are dense semantic features and depth embeddings are separated structural features, the former are easier to distinguish visually, and thus complementary information between the two modalities is achieved by predicting soft labels of RGB embedding  that can guide the latter. In addition, we argue that the ViT-based PVT network \cite{wang2021pyramid,wang2022pvt} with pyramids is stronger in feature extraction. To take advantage of this advantage, we first map PVT embedding to the same space as the image embeddings using a bias correction module, and then regard the PVT embeddings and RGB embeddings as teacher and student respectively to perform inter-model knowledge distillation, further enhancing the representation of the RGB embedding.

In brief, our contributions are summarized as follows:
\begin{itemize}
	\item We propose SAM-COD to improve SAM into the COD domain. To our knowledge, SAM-COD is the first to adopt dual stream adapters to enhance the performance of SAM on RGB-D stream based camouflaged inputs.
	\item The bidirectional distillation and mixed embedding modules are proposed. The former enriches the representations of image embeddings through model and modal distillations, and the latter updates dense prompt embeddings with hybrid image embeddings.
	\item On the four COD benchmarks, SAM-COD is thoroughly compared to existing methods. The results show that SAM-COD outperforms SOTA for both visual foundation models and specific expert models.
\end{itemize}

\section{Related Work}
\textbf{Camouflaged Object Detection (COD)} is a challenging task aimed at recognizing objects that blend in with their surroundings, and its applications cover a wide range of fields such as medicine, agriculture, and art \cite{iccv2023sam}. Traditional COD research has relied on low-level features such as texture, brightness, and color, by which foreground and background are distinguished \cite{feng2013camouflage,hou2011detection,pike2018quantifying,sengottuvelan2008performance}. However, with the advancement of deep neural networks, the development of COD has been significantly boosted. Recent studies have shown that impressive results have been achieved using methods such as mixed-scale semantics \cite{pang2022zoom}, iterative refinement \cite{jia2022segment}, and human attention mechanisms \cite{wang2021dual}. Some methods based on the design of animal hunting models, such as SINetV2 \cite{fan2021concealed} and SLSR \cite{lv2021simultaneously}, as well as methods combining probabilistic representation models with Transformer \cite{yang2021uncertainty}, have also emerged. Some methods \cite{wang2022learning,lee2022spsn,wu2023source} transformed the RGB COD task into an RGB-D COD task, attempting to utilize depth information to aid in detection. Recent studies, Bi \textit{et al.} \cite{bi2024depth} and Liu \textit{et al.} \cite{liu2024depth} have introduced dynamic allocation mechanisms and multi-scale fusion methods to suppress the interference of inaccurate depth images on COD. Different from auxiliary depth input, in our study, we learn data representation from depth images through VFM and interact RGB and depth information in an orderly manner to compensate for the unstable presentation of objects in their respective modalities.
\begin{figure*}[t]
	\centering
	\includegraphics[width=0.9\textwidth]{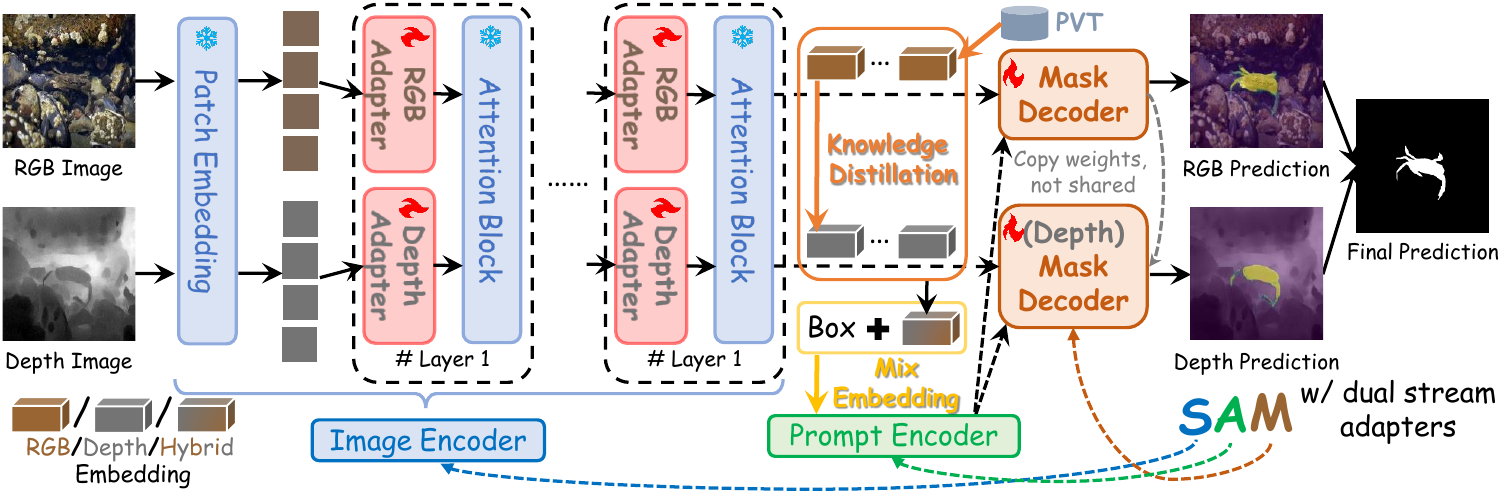}
	\vspace{-10pt}
	\caption{Overall pipeline of our SAM-COD. The dual stream images are fed into SAM in parallel to extract image features that are fine-tuned by the respective adapters. The knowledge distiller is used to address the differences caused by the dual-stream features being decoded without direct interaction. The initialized box prompt is mixed with the image features to generate a more refined dense prompt embedding. Finally, the prediction results of the two classes of feature maps are weighted summed to obtain the final detection results.} 
	\vspace{-10pt}
	\label{fig2}
\end{figure*}

\noindent\textbf{Segment Anything Model (SAM) for COD} is not consistently effective, mainly due to the large domain differences between the training data of SAM and domain-specific data, resulting in their lack of sufficient domain-specific semantic guidance. This flaw also occurs in the upgraded SAM 2 \cite{ravi2024sam} with a memory, making it difficult to generalize historical information from COD videos. For this reason, medical imaging (MI) researchers have developed models such as MedSAM-(2) \cite{ma2024segment,zhu2024medical}, Med-SA \cite{wu2023medical}, and MA-SAM \cite{chen2024ma} for tumor and myocardial segmentation. COD researchers have proposed models such as SAM(2)-Adapter \cite{iccv2023sam,chen2024sam2}, TSP-SAM \cite{hui2024endow}, and DSAM \cite{yu2024exploring} for camouflage and concealed scene recognition. SAM-COD \cite{chen2025sam,chen2024sam} with the same name as our work, which learns prior knowledge from a expert model trained on COD data via a semantic matcher and proposes an adaptive knowledge distillation strategy to ensure reliable representations, integrates three types of labels for weakly-supervised COD. Anyway, introducing domain-specific knowledge into SAM is an effective measure to improve performance and robustness on specific tasks.

Attributed to its origin, it is a feasible approach to fine-tune large pre-trained models using scalable adapters. Recently, ViT-Adapter \cite{chen2022vision} utilized adapters to enable ordinary ViTs to perform a variety of downstream tasks. Liu \textit{et al.} \cite{liu2023explicit} introduced an explicit visual prompting (EVP) technique that merges explicit visual cues into adapters. Unlike previous usage scenarios and approaches, we delve deeper into this idea by introducing dual stream adapters for RGB-D inputs to fully utilize the complementary nature of the two types of images when fine-tuning the SAM in order to facilitate the stable performance of SAM on the COD task.

\section{Method}
Figure \ref{fig2} shows the overall pipeline of the proposed SAM-COD. The dual stream adapters are embedded in the frozen SAM model to learn specific prior knowledge. Sequentially, we first briefly outline the deconstruction of the original SAM, and then introduce the technical implementation details of the modifications to the image encoder, prompt encoder, and mask decoder respectively.

\subsection{SAM Overview}
SAM \cite{kirillov2023segment} consists of an image encoder, a prompt encoder, and a mask decoder. The image encoder converts raw images into patched image embeddings using the Vision Transformer (ViT) network \cite{dosovitskiy2020image}. The prompt encoder encodes prompts (points, boxes, masks, etc.) into prompt embeddings. The mask decoder includes prompt self-attention blocks and bidirectional cross-attention blocks (prompt-to-image and reverse attentions). After applying the attention blocks, the feature maps are up-sampled and converted to segmentation masks through a fully connected layer.

In this work, we use the weights of the pre-trained SAM to initialize the weights of the three feature processors of our method, and fine-tune the mask decoder during training. According to DSAM \cite{yu2024exploring}, we input perturbed ground-truth box prompts to the original mask decoder of the SAM.

\subsection{Dual Stram Adapter}
\begin{figure}[t]
	\centering
	\includegraphics[width=\linewidth]{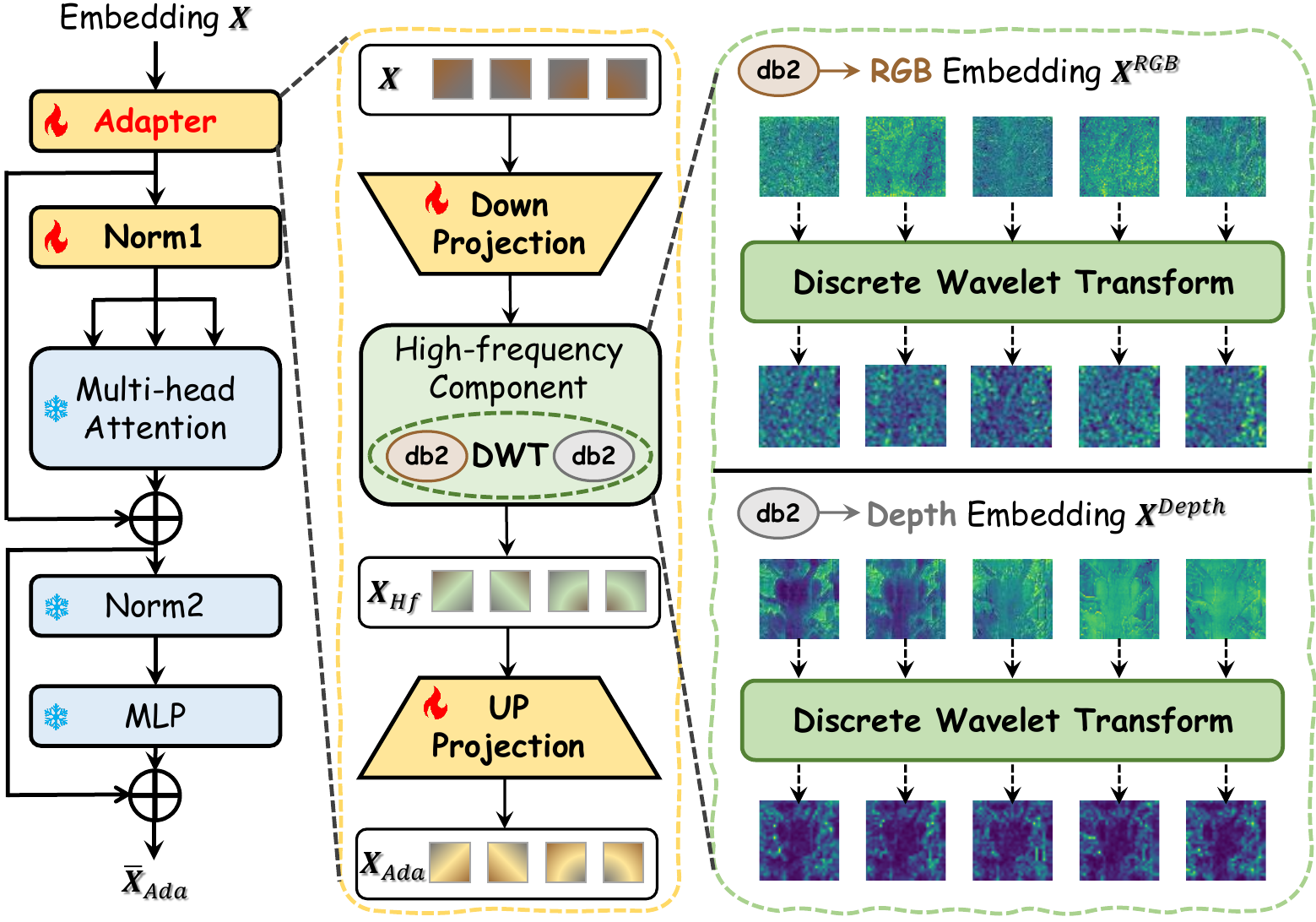}
	\vspace{-15pt}
	\caption{Illustration of the proposed adapter. The patched embedding of the image is used as input and our dual-stream adapter extracts the image embedding with high-frequency details by feature projection transform and discrete wavelet transform.}
	\vspace{-10pt}
	\label{fig3}
\end{figure}

For the input RGB image and the paired source-free depth image \cite{wu2023source}, we propose dual stream adapters for extracting high-frequency features, as shown in Figure \ref{fig3}. While keeping the both adapters simple and efficient, we further introduce different high-frequency wavelet filters for different adapters. Specifically, we choose to use an adapter consisting of two linear layers and activation functions, and the final adapter output $\bm{\bar{X}}_{Ada}$ is a combination of the input embedding $\bm{X}$ and transformed embedding by the adapter.
\begin{equation}
	\bm{\bar{X}}_{Ada}= \bm{X} + \bm{L}_{up}(\bm{\sigma}(\bm{L}_{down}(\bm{X}))),
\end{equation}
where $\bm{L}_{down}()$ is a downward projected linear layer, $\bm{\sigma}()$ is a nonlinear activation function ReLU \cite{nair2010rectified}, and $\bm{L}_{up}()$ is an upward projected linear layer. Taking the RGB or depth image embedding $\bm{X}$ as input, the computation within the adapter module obtains the image embedding with high-frequency details through the residual structure.

\noindent\textbf{DWT.} In contrast to the previous adapters that only perform downward-upward projection for single RGB inputs \cite{iccv2023sam}, our adapter is equipped with the Discrete Wavelet Transform (DWT) for extracting high-frequency features. DWT \cite{alessio2016discrete} is able to efficiently separate different frequency components of an image through multi-scale analysis, and process low-frequency and high-frequency information separately to capture subtle high-frequency features, which is particularly applicable to the changes in details of camouflaged objects, such as changes in texture and contouring. In practice, we synchronously set the same wavelet \texttt{db2} for RGB images and depth images, and merge the high-frequency subbands in the horizontal, vertical, and diagonal directions. In the case of RGB embedding,
\begin{align}
	\bm{Lf}_{RGB}, \bm{Hf}_{RGB} = DWT_{\texttt{haar}}(\bm{X}_{RGB}),\\ 
	\bm{LH}_{RGB}, \bm{HL}_{RGB}, \bm{HH}_{RGB} = \bm{Hf}_{RGB}[0|1|2],\\ 
	\bm{X}_{Hf}^{RGB} = \sqrt{\bm{LH}_{RGB}^2+\bm{HL}_{RGB}^2+\bm{HH}_{RGB}^2},
\end{align}
where $\bm{LH}$, $\bm{HL}$, and $\bm{HH}$ are the representations of the high frequency components in the three directions, respectively, and the L2 paradigm is used to synthesize the high-frequency information. By such calculation, a comprehensive high-frequency feature map can be obtained, covering the detailed variations of the image in different directions.

\noindent\textbf{Normalization.} In addition to the dual stream adapters, we also employ an activation strategy for the normalization layer \texttt{Norm1} close to the tunable adapter, which allows our SAM-COD quickly adapt to the new data distribution. At the same time, in order to prevent the pre-trained features from changing dramatically during fine-tuning, \texttt{Norm2} behind the multi-head attention layer remains frozen.

\subsection{Bidirectional Knowledge Distillation}
\begin{figure}[t]
	\centering
	\includegraphics[width=\linewidth]{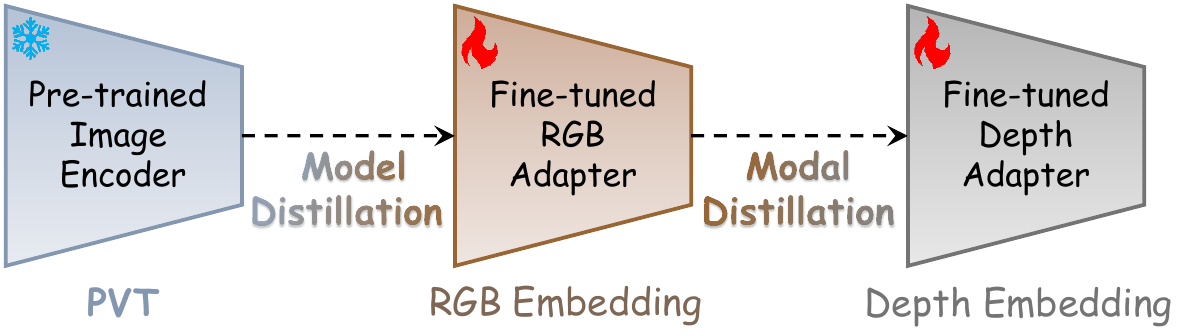}
	\vspace{-15pt}
	\caption{Illustration of the proposed bidirectional knowledge distillation. The model distillation from the pre-trained image encoder to the fine-tuned RGB adapter, and modal distillation from the RGB adapter to the depth adapter are executed sequentially.}
	\vspace{-10pt}
	\label{fig4}
\end{figure}
As illustrated in Figure \ref{fig4}, we design two knowledge distillation processes: one is the model distillation between the pre-trained encoder and the fine-tuned adapter, and the other is the modal distillation between the RGB adapter and the depth adapter. Based on the proposed bidirectional knowledge distillation (KD), the fine-tuned model not only inherits more advanced representations from the pre-trained model, but also overcomes inter-modal differences and suppresses possible noise from depth images.

\noindent\textbf{Model distillation.} Specifically, we utilize a pre-trained specialized model (\ie PVTv2 \cite{wang2022pvt}) as a teacher and pass its learned knowledge to the foundation fine-tuning model (\ie our adapter) as a student via knowledge distillation. The pre-trained model typically have stronger feature extraction capabilities and thus can provide richer learning signals to the fine-tuned model. In this way, the fine-tuning model is able to inherit the knowledge from the pre-trained model and further optimize for a specific task.
\begin{align}
	\bm{X}_{PVT}^{RGB} &= {BC}({PVT}(\bm{I})),\\
	\bm{L}_{KD-model} &= \bm{KL}(\bm{X}_{PVT}^{RGB}, \bm{\bar{X}}_{Ada}^{RGB}),
\end{align}
where $\bm{I} \in \mathbb{R}^{H\times W\times3}$ is an RGB image, ${PVT}$ is the frozen pre-trained model and ${BC}$ is the trainable bias-calibrated network ${BC}(\bm{X}) = {\alpha}*\bm{X}+{\beta}$ that maps $\bm{X}$ to the dimensions of $\bm{\bar{X}}_{Ada}^{RGB}$. $\bm{KL}(Teacher,Student)$ is the Kullback-Leibler divergence \cite{kullback1951information}, which is used to measure the difference between the teacher model and the student model.

\noindent\textbf{Modal distillation.} While $\bm{X}_{Hf}^{RGB}$ guided by a specialized model is rich in semantic information, $\bm{X}_{Hf}^{Depth}$ consists of deep structural information that usually contains certain noise and does not always highlight the camouflaged object \cite{lv2023toward}. Therefore, we continue to perform modal distillation by using RGB embeddings as teachers to instruct depth embeddings as students to learn the semantic knowledge.
\begin{equation}
	\bm{L}_{KD-modal} = \bm{KL}(\bm{\bar{X}}_{Ada}^{RGB}, \bm{\bar{X}}_{Ada}^{Depth}).
\end{equation}

As a result, we utilize the KD metrics $\bm{L}_{KD-model}$ and $\bm{L}_{KD-modal}$ of the bidirectional distillation as loss items during the training phase. This strategy can enhance the representation of the modified visual foundation model to improve robustness and accuracy in the COD task.

\subsection{Mixed Prompt Embedding}
During image encoding, SAM causes loss of detail information due to 16×16 downsampling. To solve this issue, in addition to using dense PVT embedding to guide image embedding, we additionally use dual stream embeddings to compensate for the dense prompt embedding.

Specifically, first the box prompt embedding is processed by a deepwise convolutional layer; then the expert and adapter embeddings are concatenated to generate the new hybrid embedding; finally the dense prompt embedding is updated by another deepwise convolutional layer. This process can effectively fuse the dual stream image and prompt embeddings to enhance the representation capability of new dense prompt embedding. Mathematically,
\begin{align}
	\bm{X}_{Hybrid} &= cat[\bm{X}_{PVT}^{RGB}, \bm{\bar{X}}_{Ada}^{RGB}+ \bm{\bar{X}}_{Ada}^{Depth}],\\
	\bm{X}_{box}^{dense} &= DWConv(cat[\bm{X}_{box}^{dense}, \bm{X}_{Hybrid}]),
\end{align}
where $cat[*,*]$ represents the embedding concatenation operation. Here $DWConv$ is used for high-frequency filtering in the diagonal direction, as we observe that it applies to the summation of dual stream embeddings.

\subsection{Training and Inference}
In the training phase, we optimize our model using two losses. On the one hand, DiceCELoss is employed as a loss function for dual stream predictions, which computes the weighted sum of the dice loss and the cross-entropy loss. On the other hand, a bidirectional distillation loss is employed to utilize the in-channel knowledge to achieve soft alignment between the model and the modal. In summary, the loss function of our SAM-COD is as follows.
\begin{align}
	\bm{L}_{DiceCE} &= DiceCE(\bm{Y}^{RGB}|\bm{Y}^{Depth}, \bm{Y}^{GT})\\
	\bm{L} &= \lambda\bm{L}_{DiceCE}  + (1-\lambda)\bm{L}_{KD},
\end{align}
where $\bm{Y}^{GT}$ denotes the binarized ground truth and is used twice to supervises dual stream predictions $\bm{Y}^{RGB}$ and $\bm{Y}^{Depth}$ respectively. Moreover, $\lambda$ is a hyperparameter used to balance the prediction loss and distillation loss of SAM-COD, and the default value is 0.9.

In order to obtain the final detection result during inference, we weight the RGB prediction $\bm{Y}^{RGB}$ and depth prediction $\bm{Y}^{Depth}$. In the end, pixels with a mean value greater than 0.5 are considered to belong to the foreground pixel, otherwise they are considered to be background pixels.
\begin{equation}
	\bm{Y} = (0.5 * \bm{Y}^{RGB} + 0.5 * \bm{Y}^{Depth}) > 0.5.
\end{equation}

Considering the real application where depth images may not be available or of poor quality, we can use only the RGB stream loss and prediction as a result of our SAM-COD. The construction and results of this setting will be discussed in the subsequent experimental analysis.

\section{Experiments}
\subsection{Experimental Settings}
\noindent\textbf{Datasets.} The experiments are conducted on four datasets, namely CHAMELEON \cite{skurowski2018animal}, CAMO \cite{le2019anabranch}, COD10K \cite{fan2021concealed} and NC4K \cite{lv2021simultaneously}. CHAMELEON contains 76 images collected from the Internet for testing. CAMO contains 1250 images randomly divided into 1000 training and 250 test sets. COD10K contains 5066 images, of which 3040 are training and 2026 are test sets. NC4K contains 4121 images, all of which are used as test sets. Following protocol from \cite{fan2021concealed}, our study uses a dataset consisting of the training set of COD10K and CAMO, which contains 3040 images and 1000 images, respectively, and the rest of the images of COD10K and CAMO and the entire CHAMELEON and NC4K images are used as the test dataset. 

\noindent\textbf{Evaluation metrics.} The six evaluation metrics in the COD field are used, including the structural metric $S_\alpha$ \cite{fan2017structure}, the adaptive/maximum $E$-measure $\alpha E$/$E^x$ \cite{fan2018enhanced}, the weighted/maximum $F$-measure $F^{\omega}$/$F^x$ \cite{margolin2014evaluate},  and average absolute error $M$. Where, $S_\alpha$ measures the structural similarity between the predicted results and the actual segmented region. $E$-measure considers the structure and texture difference of two images. $F$-measure is the harmonic mean of precision and recall. $M$ is the average absolute difference between the predicted value and the true value.

\noindent\textbf{Implementation details.} We implement our SAM-COD in Pytorch \cite{paszke2019pytorch}. The pre-trained models consist of the specialized weight PVTv2 \cite{wang2022pvt} and VFM weight SAM-B \cite{kirillov2023segment}. We resize the inputs of all images to 1024x1024 by bilinear interpolation, train on a NVIDIA 3090 GPU with the batch size of 1, and use the AdamW optimizer \cite{loshchilov2017fixing} with an initialized learning rate of $1\times10^{-4}$ for 60 epochs.

\subsection{Comparison Results}
\begin{table*}[t]
	\caption{Quantitative results of different COD methods on four datasets, \ie, CAMO \cite{le2019anabranch},  CHAMELEON \cite{skurowski2018animal}, COD10K \cite{fan2021concealed} and NC4K \cite{lv2021simultaneously}. \textcolor{red}{Red}/\textcolor{blue}{blue}/\textcolor{orange}{orange} indicates the \textcolor{red}{1st}/\textcolor{blue}{2nd}/\textcolor{orange}{3rd} best result for the current setting. $\dag$ represents the SAM-based methods.}
	\vspace{-10pt}
	\label{tab1}
	\centering
	\setlength{\tabcolsep}{0.4mm}
	\resizebox{\linewidth}{!}{
	\begin{tabular}{l|cccccc|cccccc|cccccc|cccccc}
		\toprule 
		\multirow{2}*{Method} & \multicolumn{6}{c|}{CAMO (250)} & \multicolumn{6}{ c|}{CHAMELEON (76)} & \multicolumn{6}{ c|}{COD10K (2026)} & \multicolumn{6}{ c}{NC4K (4121)} \\
		& $M\downarrow$ & $F^x\uparrow$ & $F^{\omega}\uparrow$ & $S_m\uparrow$ & $E^x\uparrow$ & $\alpha E\uparrow$ & 
		$M\downarrow$ & $F^x\uparrow$ & $F^{\omega}\uparrow$ & $S_m\uparrow$ & $E^x\uparrow$ & $\alpha E\uparrow$ & 
		$M\downarrow$ & $F^x\uparrow$ & $F^{\omega}\uparrow$ & $S_m\uparrow$ & $E^x\uparrow$ & $\alpha E\uparrow$ & 
		$M\downarrow$ & $F^x\uparrow$ & $F^{\omega}\uparrow$ & $S_m\uparrow$ & $E^x\uparrow$ & $\alpha E\uparrow$ \\
		\midrule 
		\multicolumn{25}{c}{RGB-based COD Methods} \\
		\cdashline{1-25}[4pt/4pt]
		SINet \cite{fan2020camouflaged} & .099 & .762 & .606 & .751 & .790 & .825 & 
		.044 & .845 & .740 & .868 & .908 & .938 & 
		.051 & .708 & .551 & .771 & .832 & .867 & 
		.058 & .804 & .723 & .808 & .873 & .883 \\
		SLSR \cite{lv2021simultaneously} & .080 & .791 & .696 & .787 & .843 & .855 &  .030 & .866 & - & .889 & \color{orange}.938 & - & 
		.037 & .756 & .673 & .804 & .854 & .882 & 
		.048 & .836 & .766 & .839 & .898 & .902 \\
		MGL-R \cite{zhai2021mutual} & .088 & .791 & .719 & .775 & .820 & .848 & 
		.031 & .868 & .828 & .893 & .932 & .923 & 
		.035 & .767 & .686 & .813 & .874 & .865 & 
		.053 & .828 & .762 & .832 & .876 & .867 \\
		PFNet \cite{mei2021camouflaged} & .085 & .793 & .695 & .782 & .845 & .852 & 
		.033 & .859 & .810 & .882 & .927 & .942 & 
		.040 & .747 & .660 & .800 & .880 & .868 & 
		.053 & .820 & .745 & .829 & .891 & .894 \\
		UJSC \cite{li2021uncertainty} & .072 & \color{orange}.812 & .728 & .800 & .861 & .853 & 
		.030 & \color{blue}{.874} & - & .891 & \color{blue}{.948} & - & 
		.035 & .761 & .684 & .808 & .886 & .891 & 
		.047 & \color{orange}.838 & .771 & .841 & .900 & .907 \\
		C2FNet \cite{sun2021context} & .079 & .802 & .719 & .796 & .856 & .864 & 
		.032 & \color{orange}.871 & .828 & .888 & .936 & .932 & 
		.036 & .764 & .686 & .813 & .894 & .886 & 
		.049 & .831 & .762 & .838 & .898 & .901 \\
		UGTR \cite{yang2021uncertainty} & .086 & .800 & .695 & .783 & .829 & .859 & 
		.031 & .862 & .810 & .887 & .926 & .921 & 
		.036 & \color{orange}.769 & .660 & .816 & .873 & .850 & 
		.052 & .831 & .745 & .839 & .884 & .889 \\
		SegMAR \cite{jia2022segment} & .080 & .799 & .742 & .794 & .857 & .872 & 
		.032 & \color{orange}.871 & .835 & .887 & .935 & .950 & 
		.039 & .750 & .724 & .799 & .876 & .895 & 
		.050 & .828 & .781 & .836 & .893 & .905 \\
		ZoomNet \cite{pang2022zoom} & {.074} & \color{blue}{.818} & {.752} & .801 & .858 & .883 & 
		.033 & .829 & \color{orange}{.845} & .859 & .915 & .952 & 
		.034 & \color{blue}.771 & .729 & .808 & .872 & .893 & 
		.045 & \color{blue}{.841} & .784 & .843 & .893 & .907 \\
		SINetv2 \cite{fan2021concealed} & .070 & - & .743 & .820 & .882 & .875 & 
		.030 & - & .816 & .888 & - & .942 & 
		.037 & - & .680 & .815 & .887 & .863 &
		.048 & - & .770 & .847 & .903 & .898  \\
		DGNet \cite{ji2023deep} & \color{orange}{.057} & - & .769 & .839 & \color{orange}{.915} & \color{blue}.901 & 
		{.029} & - & .816 & .890 & - & .934 & 
		.033 & - & .693 & .822 & .911 & .877 &
		.042 & - & .784 & .857 & .922 & .907 \\
		FSPNet \cite{huang2023feature} & \color{blue}{.050} & - & \color{blue}{.799} & \color{blue}{.856} & \color{blue}{.928} & \color{orange}.899 & 
		- & - & - & - & - & - & 
		\color{orange}{.026} & - & .735 & .851 & \color{blue}{.930} & .895 & 
		.035 & - & \color{orange}.816 & \color{blue}.879 & \color{blue}.937 & \color{orange}.915 \\
		CamoFormer-R \cite{yin2024camoformer} & .066 & - & .756 & .817 & - & .884 & 
		\color{blue}{.024} & - & .843 & \color{blue}.900 & - & .949 & 
		.029 & - & .730 & .838 & - & .898 & 
		\color{red}.024 & - & .793 & .857 & - & \color{orange}.915\\
		CamoFocus-R \cite{khan2024camofocus} & .071 & - & .752 & .812 & - & .873 & 
		\color{orange}.027 & - & \color{blue}{.849} & \color{orange}.898 & - & \color{orange}{.953} & 
		.033 & - & .719 & .825 & - & \color{orange}.903 & 
		.043 & - & .788 & .847 & - & .910\\
		ZoomNeXt-R \cite{pang2024zoomnext} & .069 & - & .760 & .822 & - & .885 & 
		\color{red}{.020} & - & \color{red}{.864} & \color{red}{.912} & - & \color{red}{.969} & 
		\color{orange}{.026} & - & \color{orange}{.758} & \color{orange}{.855} & - & \color{blue}{.926} & 
		\color{orange}.038 & - & .808 & \color{orange}.869 & - & \color{blue}.925\\
		SAM$^\dag$ \cite{kirillov2023segment} & .132 & - & .606 & .684 & .689 & - & 
		- & - & - & - & - & - & 
		.049 & - & .701 & .783 & .800 & - & 
		.078 & - & .696 & .767 & .778 & -\\
		SAM-Adapter$^\dag$ \cite{iccv2023sam} & .070 & - & .765 & \color{orange}.847 & .873 & - & 
		.033 & - & .824 & .896 & .919 & - & 
		\color{blue}{.025} & - & \color{blue}{.801} & \color{red}{.883} & .918 & - & 
		- & - & - & - & - & -\\
		MedSAM$^\dag$ \cite{ma2024segment} & .065 & - & \color{orange}.780 & .820 & .913 & - & 
		- & - & - & - & - & - & 
		.033 & - & .751 & .841 & \color{orange}.926 & - & 
		.041 & - & \color{blue}{.821} & .866 & \color{blue}.937 & -\\
		\rowcolor{gray!15}\textbf{SAM-COD$^\dag$} & \color{red}{.047} & \color{red}{.876} & \color{red}{.839} & \color{red}{.866} & \color{red}{.948} & \color{red}{.946}
		& {.031} & \color{red}{.881} & {.841} & {.888} & \color{red}{.965} & \color{blue}{.956}
		& \color{red}{.023} & \color{red}{.866} & \color{red}{.817} & \color{blue}{.881} & \color{red}{.969} & \color{red}{.942}
		& \color{blue}{.032} & \color{red}{.892} & \color{red}{.857} & \color{red}{.889} & \color{red}{.963} & \color{red}{.954} \\
		\midrule
		\multicolumn{25}{c}{RGB-D-based COD Methods} \\
		\cdashline{1-25}[4pt/4pt]
		CDINet \cite{zhang2021cross} & .100 & .638 & - & .732 & .766 & - & 
		.036 & .787 & - & .879 & .903 & - & 
		.044 & .610 & - & .778 & .821 & - & 
		.067 & .697 & - & .793 & .830 & - \\
		DCF \cite{ji2021calibrated} & .089 & .724 & - & .749 & .834 & - & 
		.037 & .821 & - & .850 & .923 & - & 
		.040 & .685 & - & .766 & .864 & - & 
		.061 & .765 & - & .791 & .878 & - \\
		CMINet \cite{zhang2021rgb} & .087 & .798 & - & .782 & .827 & - & 
		\color{orange}.032 & \color{orange}{.881} & - & \color{orange}{.891} & .930 & - & 
		.039 & .768 & - & .811 & .868 & - & 
		.053 & .832 & - & .839 & .888 & - \\
		SPNet \cite{zhou2021specificity} & .083 & .807 & - & .783 & .831 & - & 
		.033 & .872 & - & .888 & .930 & - & 
		.037 & .776 & - & .808 & .869 & - & 
		.054 & .828 & - & .825 & .874 & - \\
		DCMF \cite{wang2022learning} & .115 & .737 & - & .728 & .757 & - & 
		.059 & .807 & - & .830 & .853 & - & 
		.063 & .679 & - & .748 & .776 & - & 
		.077 & .782 & - & .794 & .820 & - \\
		SPSN \cite{lee2022spsn} & .084 & .782 & - & .773 & .829 & - & 
		\color{orange}.032 & .866 & - & .887 & .932 & - & 
		.042 & .727 & - & .789 & .854 & - & 
		.059 & .803 & - & .813 & .867 & - \\
		PopNet \cite{wu2023source} & \color{orange}{.073} & \color{orange}{.821} &  & .806 & \color{orange}.869 &  & 
		\color{red}{.022} & \color{red}{.893} & - & \color{red}{.910} & \color{blue}{.962} & - & 
		\color{blue}.031 & \color{orange}.789 & - & \color{orange}.827 & \color{orange}.897 & - & 
		\color{orange}.043 & \color{orange}.852 & - & \color{orange}.852 & \color{orange}.908 & - \\
		DSAM$^\dag$ \cite{yu2024exploring} & \color{blue}{.061} & \color{blue}{.834} & \color{orange}{.794} & \color{blue}{.832} & \color{blue}{.920} & \color{orange}{.920} & 
		.042 & .824 & \color{orange}{.784} & .854 & \color{orange}{.935} & \color{blue}{.925} & 
		\color{orange}.033 & \color{blue}.807 & \color{blue}{.758} & \color{blue}{.846} & \color{blue}{.931} & \color{blue}{.912} & 
		\color{blue}.040 & \color{blue}{.862} & \color{blue}{.828} & \color{blue}{.871} & \color{blue}{.940} & \color{blue}{.936} \\
		\rowcolor{gray!15}\textbf{SAM-COD$^\dag$} & \color{red}{.044} & \color{red}{.884} & \color{red}{.849} & \color{red}{.875} & \color{red}{.953} & \color{red}{.952}
		& \color{blue}{.029} & \color{blue}{.887} & \color{red}{.850} & \color{blue}{.893} & \color{red}{.969} & \color{red}{.961}
		& \color{red}{.022} & \color{red}{.873} & \color{red}{.827} & \color{red}{.887} & \color{red}{.972} & \color{red}{.948}
		& \color{red}{.029} & \color{red}{.898} & \color{red}{.866} & \color{red}{.896} & \color{red}{.972} & \color{red}{.959} \\
		\bottomrule
		\end{tabular}
	}
	\vspace{-10pt}
\end{table*}


We compare our SAM-COD with 18 RGB stream-based, and 8 RGB-D stream-based state-of-the-art (SOTA) COD methods which are converted from the salient object detection (SOD) methods CDINet \cite{zhang2021cross}, DCF \cite{ji2021calibrated}, CMINet \cite{zhang2021rgb}, SPNet \cite{zhou2021specificity}, DCMF \cite{wang2022learning}, SPSN \cite{lee2022spsn}, PopNet \cite{wu2023source}. In particular, we focus on the performance of VFM methods with the SAM weight, including SAM \cite{kirillov2023segment}, SAM-Adapter \cite{iccv2023sam}, MedSAM \cite{ma2024segment} and DSAM \cite{yu2024exploring}. 

Among them, although MedSAM \cite{ma2024segment} was initially applied in the field of medical image processing, it belongs to the enhanced SAM, and its application in COD also significantly improves the performance. In addition, we abandon the SAM-Adapter \cite{iccv2023sam} using SAM-H due to the conditions. For fairness, prediction results are provided directly from their papers or generated by their already trained models.

\noindent\textbf{Quantitative evaluation.} Table \ref{tab1} shows that the proposed SAM-COD outperforms other SOTA methods on most of the evaluation metrics in b\underline{}oth settings. On the one hand, when our SAM-COD is oriented only to the RGB stream, it outperforms the existing methods substantially, bringing an overall improvement of 2\%-6\% on four datasets and breaking through the bottleneck of existing technologies. Moreover, we notice that compared to recent CNN-based or Transformer-based COD methods (e.g., ZoomNet \cite{pang2022zoom}, SegMaR \cite{jia2022segment}, and CamoFormer \cite{yin2024camoformer}), although they employ multi-stage or multi resolution training and inference and other strategies to enhance model performance which add additional computational burden, our SAM-COD constructed based on the VFM model (SAM-B) still outperforms them in most benchmark tests. On the other hand, when SAM-COD is oriented to the RGB-D stream, \ie, with the aid of source-free depth images, it is more capable of mining the detailed information of camouflaged objects.
\begin{figure*}[t]
	\centering
	\includegraphics[width=\textwidth]{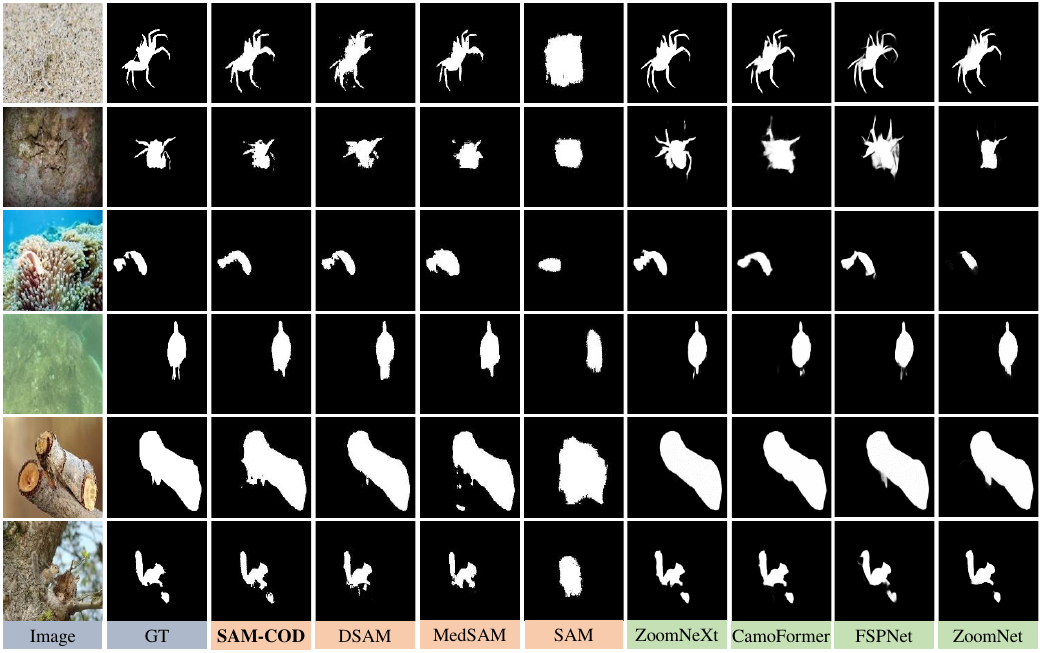}
	\vspace{-20pt}
	\caption{Comparison of our SAM-COD and other methods in the COD task. We are mainly concerned with those of SAM-based methods.} 
	\vspace{-15pt}
	\label{fig5}
\end{figure*}

\noindent\textbf{Qualitative evaluation.} We perform qualitative visualization results of SAM-COD on eight scenes selected from four datasets to compare with three methods using SAM, as show in Figure \ref{fig5}. We analyze four aspects: object size, edge complexity, perceived missing parts, and multiple objects. It can be found that our SAM-COD can effectively balance contours and details, and can achieve clear segmentation of edges for objects with high edge complexity (fourth column). In addition, SAM-COD can correctly segment highly camouflaged parts and extra objects (the first and eighth columns) that are easily overlooked, on the contrary, other methods may exhibit omission or over-segmentation. In short, SAM techniques themselves lack a comprehensive understanding of the foreground and background, making them misclassify certain regions or miss some objects when dealing with complex conditions. In contrast, our SAM-COD corrects this information to a great extent with the dual stream adapters.
\begin{figure}[t]
	\centering
	\includegraphics[width=\linewidth]{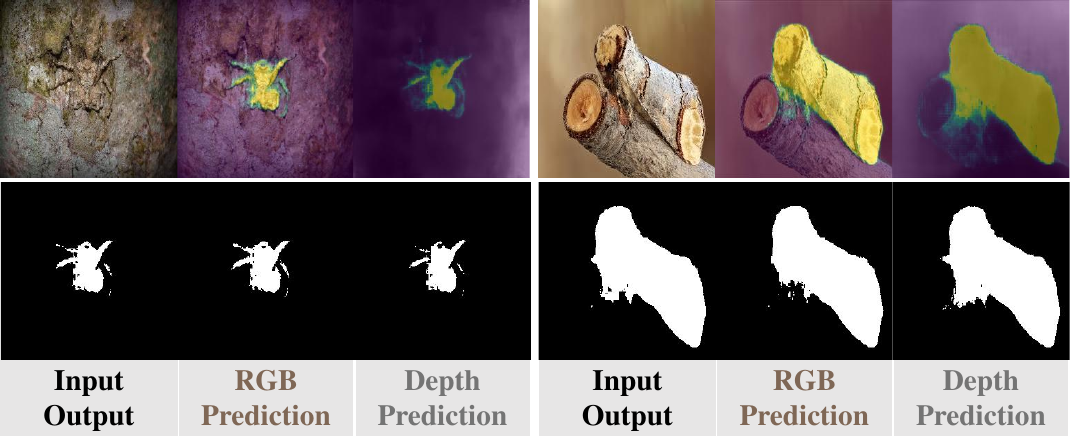}
	\vspace{-15pt}
	\caption{Comparison of the dual-stream adapter for dual-stream inputs. The final prediction result is optimized by collecting common high-confidence regions from the dual stream predictions.}
	\vspace{-15pt}
	\label{fig6}
\end{figure}

\noindent\textbf{Visualization evaluation.} To show more clearly the impact of our dual-stream adapters on the dual-stream inputs, we visualize their respective feature maps and detection results after binarization (corresponding to Figure \ref{fig5}), and the final detection results, as shown in Figure \ref{fig6}. It can be seen that the dual-stream adapters are able to capture their respective details from different graphs, which complement each other to obtain predictions that ultimately optimize both. More results can be found in the supplementary material.
\begin{figure}[t]
	\centering
	\includegraphics[width=\linewidth]{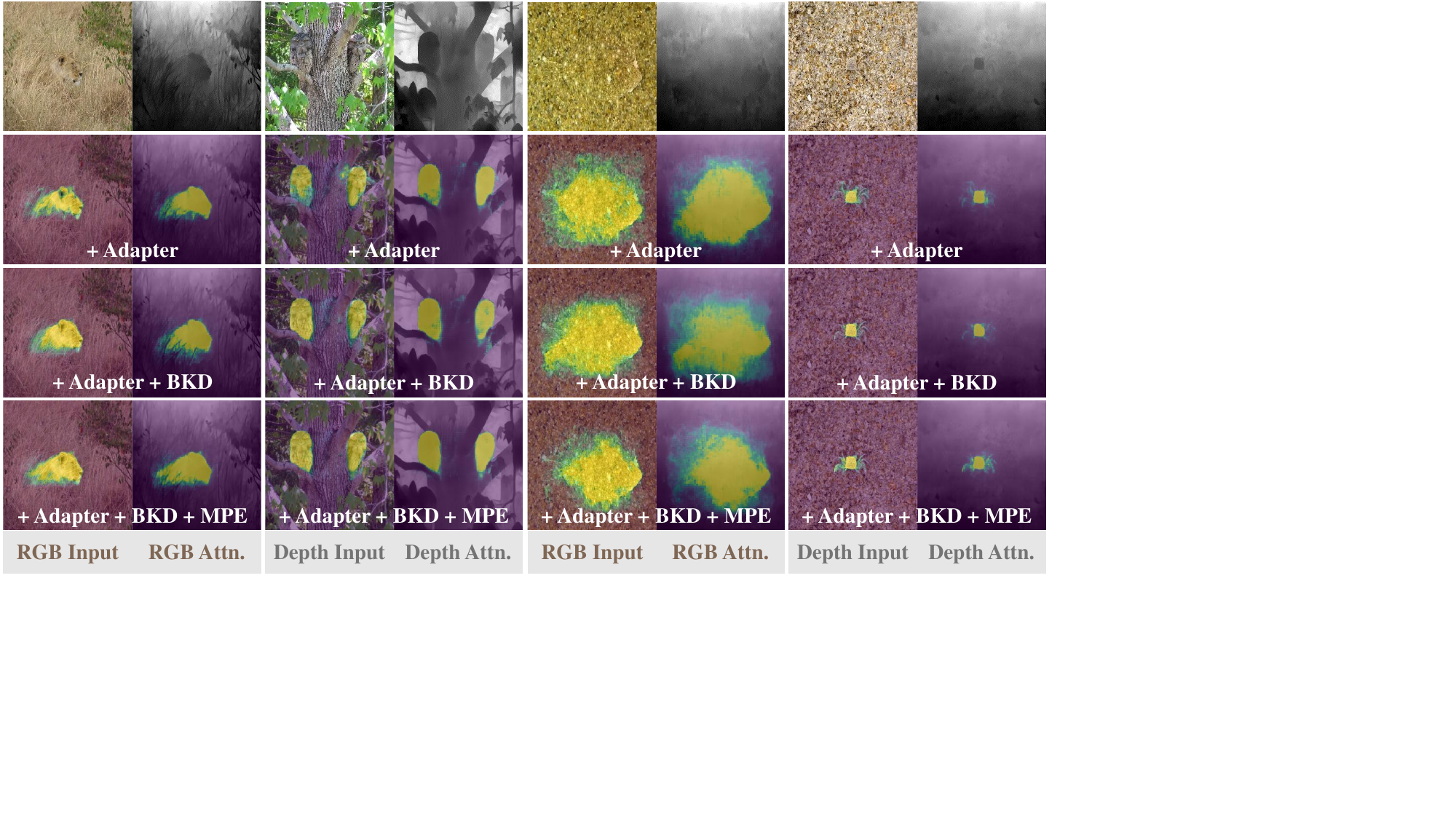}
	\vspace{-15pt}
	\caption{Attentional visualization of dual-stream inputs with Adapter, BKD, and MPE modules, respectively. The components used for ablation study increase cumulatively from top to bottom.}
	\vspace{-15pt}
	\label{fig7}
\end{figure}

\subsection{Ablation Study}
To demonstrate the impact of each module from SAM-COD, we conduct ablation studies by selectively adding and removing modules from the baseline. Consequently, we consider replacing existing components and discover more experimental phenomena to confirm the reliability of the our method, and the results are shown in Table \ref{tab2}.

\textbf{Component ablation.} Our SAM-COD consists of three main components: the core dual-stream adapter (Adapter) module, and the bidirectional knowledge distillation (BKD) and mixed prompt embedding (MPE) modules to better adapt to dual-stream representation learning. The effect of Adapter on the baseline is obvious, with an average improvement of about 8\% on the positive evaluation criteria on COD10K. Note that BKD and MPE have a more discriminative effect on the learning of dual-stream inputs, especially in the details of the fusion of the target and the environment. To show this phenomenon intuitively, we show the mask prediction of their application in Figure \ref{fig7}. It can be found that the combination of BKD and MPE can make the adapter more accurate in perceiving camouflaged objects.
\begin{table*}[t]
	\caption{Ablation study and performance comparison of each module on four benchmark datasets.}
	\vspace{-10pt}
	\label{tab2}
	\centering
	\setlength{\tabcolsep}{1.4mm}
	\resizebox{\linewidth}{!}{
		\begin{tabular}{l|cccc|cccc|cccc|cccc}
			\toprule 
			\multirow{2}*{Ablation settings} & \multicolumn{4}{c|}{CAMO (250)} & \multicolumn{4}{c|}{CHAMELEON (76) } & \multicolumn{4}{ c|}{COD10K (2026)} & \multicolumn{4}{c}{NC4K (4121)} \\
			& $M\downarrow$ & $F^{\omega}\uparrow$ & $S_m\uparrow$ & $\alpha E\uparrow$ & 
			$M\downarrow$ & $F^{\omega}\uparrow$ & $S_m\uparrow$ & $\alpha E\uparrow$ & 
			$M\downarrow$ & $F^{\omega}\uparrow$ & $S_m\uparrow$ & $\alpha E\uparrow$ & 
			$M\downarrow$ & $F^{\omega}\uparrow$ & $S_m\uparrow$ & $\alpha E\uparrow$ \\
			\midrule 
			\multicolumn{17}{c}{Ablation study} \\
			\cdashline{1-17}[4pt/4pt]
			Baseline (B) & .073 & .759 & .803 & .901
			& .054 & .729 & .815 & .915
			& .037 & .735 & .829 & .908
			& .047 & .804 & .854 & .925 \\
			B + Adapter & .056 & .814 & .849 & .935
			& .034 & .827 & .875 & .953
			& .026 & .794 & .865 & .938
			& .035 & .846 & .881 & .951 \\
			B + Adapter + BKD & .051 & .823 & .863 & .945
			& .032 & .834 & .882 & .949
			& .023 & .811 & .870 & .940
			& .032 & .854 & .887 & .953 \\
			\rowcolor{gray!15}B + Adapter + BKD + MPE & .044 & .849 & .875 & .952
			& .029 & .850 & .893 & .961
			& .022 & .827 & .887 & .948
			& .029 & .866 & .896 & .959 \\
			\midrule
			\multicolumn{17}{c}{Adapter structure} \\
			\cdashline{1-17}[4pt/4pt]
			w/ LoRA form & .056 & .815 & .847 & .936
			& .034 & .819 & .870 & .955
			& .029 & .769 & .848 & .933
			& .041 & .816 & .860 & .941 \\
			w/o DWT & .050 & .834 & .860 & .944
			& .031 & .842 & .885 & .962
			& .025 & .803 & .868 & .944
			& .036 & .840 & .875 & .951 \\
			RGB-stream adapter & .047 & .839 & .866 & .946
			& .031 & .841 & .888 & .956
			& .023 & .817 & .881 & .942
			& .032 & .857 & .889 & .954 \\
			\rowcolor{gray!15}Dual-stream adapter & .044 & .849 & .875 & .952
			& .029 & .850 & .893 & .961
			& .022 & .827 & .887 & .948
			& .029 & .866 & .896 & .959 \\
			\midrule
			\multicolumn{17}{c}{Knowledge distillation} \\
			\cdashline{1-17}[4pt/4pt]
			$\bm{L}_{KD-model} (\bm{L}_{PVT \rightarrow RGB})$ & .048 & .838 & .865 & .945
			& .031 & .844 & .888 & .960
			& .024 & .812 & .875 & .947
			& .035 & .842 & .878 & .951 \\
			$\bm{L}_{KD-modal} (\bm{L}_{RGB \rightarrow Depth})$ & .048 & .840 & .867 & .946
			& .030 & .846 & .889 & .961
			& .024 & .815 & .876 & .948
			& .035 & .843 & .879 & .953 \\
			$\bm{L}_{(PVT \rightarrow Depth)} $ + $\bm{L}_{(Depth \rightarrow RGB)} $ & .047 & .842 & .868 & .948
			& .029 & .847 & .890 & .962
			& .023 & .818 & .880 & .948
			& .033 & .848 & .882 & .957 \\
			\rowcolor{gray!15}$\bm{L}_{(PVT \rightarrow RGB)} $ + $\bm{L}_{(RGB \rightarrow Depth)} $ & .044 & .849 & .875 & .952
			& .029 & .850 & .893 & .961
			& .022 & .827 & .887 & .948
			& .029 & .866 & .896 & .959 \\
			\midrule
			\multicolumn{17}{c}{Prompt embedding} \\
			\cdashline{1-17}[4pt/4pt]
			$\bm{\bar{X}}_{Ada}^{RGB} or \bm{\bar{X}}_{Ada}^{Depth}$ & .055 & .811 & .848 & .934
			& .037 & .810 & .865 & .947
			& .029 & .771 & .852 & .930
			& .040 & .817 & .863 & .941 \\
			$\bm{\bar{X}}_{Ada}^{RGB} + \bm{\bar{X}}_{Ada}^{Depth}$ & .053 & .819 & .851 & .939
			& .036 & .814 & .866 & .950
			& .029 & .776 & .853 & .932
			& .038 & .829 & .868 & .945 \\
			$cat[\bm{X}_{Ada}^{RGB}, \bm{\bar{X}}_{Ada}^{Depth}]$ & .049 & .838 & .864 & .947
			& .033 & .841 & .884 & .961
			& .024 & .812 & .873 & .948
			& .035 & .843 & .878 & .952 \\
			\rowcolor{gray!15}$cat[\bm{X}_{PVT}^{RGB}, \bm{\bar{X}}_{Ada}^{RGB}+ \bm{\bar{X}}_{Ada}^{Depth}]$ & .044 & .849 & .875 & .952
			& .029 & .850 & .893 & .961
			& .022 & .827 & .887 & .948
			& .029 & .866 & .896 & .959 \\
			\bottomrule
		\end{tabular}
	}
	\vspace{-10pt}
\end{table*}

\textbf{Adapater ablation.} We develop a series of detailed discussions of the adapter, including adapted features added incrementally to frozen blocks of attention (LoRA \cite{hu2022lora} designed for image vision), without discrete wavelet transforms, 
and using the adapter only for RGB inputs. From the results, the adapter in the form of LoRA is not able to perceive the camouflaged object effectively with limited incremental cues. The features learned by the dual-stream adapter without DWT have certain defects. Moreover, Norm1 and Norm2 have less impact on the final prediction results, which shows that our adapter do not depend on the architecture of the original VFM network. Last but not list, for the RGB input only, which is also a realistic condition as discussed before, the detection performance is slightly degraded without the complement of depth information, but it also outperforms numerous advanced methods with the same input setting.

\textbf{Distillation ablation.} For the ablation study of knowledge distillation, we first study the effect of single knowledge distillation, that is, the knowledge transfer occurs in $\bm{L}_{KD-model}$ between models and $\bm{L}_{KD-modal}$ between modalities. From the results, we find that traditional single knowledge distillation cannot adapt to our dual-stream adapter. Due to the lack of expert knowledge prior, only $\bm{L}_{KD-modal}$ performs the worst. Then we try to use expert embedding (PVT) to guide the adapter embedding (Depth), and then establish a relationship with embedding (RGB). This order of BKD is different from ours, which we analyze: as PVT is essentially trained on a large number of RGB images, the dense PVT embedding first guides the regional depth embedding, which has a slight gap in results.

\textbf{Prompt ablation.} We utilize the prompt embedding from SAM and use sparse box embeddings in the initial stage. Then to make the prompt embeddings dense pixel-aware, we use point embeddings generated in the image encoder mixed with the prompt embeddings. The first option is to use separate adapter embeddings for different inputs, which turns out to be the worst configuration due to the presence of learning bias in the mixing process. Then we use the common feature summation operation, using the dual-stream embedding as mixed pixel features. It can be found that the embedding summation yields better results, as this operation directly integrates the representations of the two types of inputs. We then introduce the expert embedding PVT, which yields moderate gains when simply concatenated with the RGB embedding. Finally, we achieve the best results with our default configuration after adding deep embeddings to the image representation.

\section{Conclusion}\label{sec5}
This paper proposes an improved visual foundation model SAM-COD for the camouflaged object detection task. By extending the dual stream adapters, introducing the bidirectional knowledge distillation and mixed prompt embedding, the proposed SAM-COD significantly improves the performance of the segmentation performance on RGB-D input. Experimental results show that SAM-COD can effectively utilize the complementary information of RGB and depth images, provide an efficient solution for camouflaged object detection. This approach further demonstrates that significant performance improvements can be achieved by designing modules to suit specific tasks while maintaining the integrity of the infrastructure.

\noindent\textbf{Future work.} Our future work will be dedicated to further optimizing SAM-COD to handle more complex multimodal inputs, such as combining thermal imaging data or spectral data to improve the applicability of the model. In addition, the lightweight SAM and adapter can be explored to reduce computational complexity and make it more efficient in real-time applications. We also plan to verify the versatility of SAM-COD in more downstream tasks, such as small object detection or multi-instance segmentation, to fully explore its potential in 2D vision tasks.

{\small
\bibliographystyle{ieee_fullname}
\bibliography{reference}
}

\end{document}

%% file: preamble.tex
\usepackage{amsmath,amsfonts,bm}
\usepackage{algorithmic}
\usepackage{algorithm}
\usepackage{arydshln}
\usepackage{array}
\usepackage{bbding}
\usepackage{colortbl}
\usepackage{stfloats}
\usepackage{url}
\usepackage{verbatim}
\usepackage{wasysym}
\usepackage{pifont}
\usepackage{cancel} 
\usepackage{graphicx}
\usepackage{amsmath}
\usepackage{amssymb}
\usepackage{booktabs}
\usepackage{multirow}
\usepackage{bm} 
\usepackage{diagbox}

\graphicspath{{figures/}}
